\documentclass[conference]{IEEEtran}
\IEEEoverridecommandlockouts
\usepackage{cite}
\usepackage{amsmath,amssymb,amsfonts}
\usepackage[font=footnotesize]{subcaption}
\usepackage{algorithmic}
\usepackage{graphicx}
\usepackage{textcomp}
\usepackage{xcolor}
\def\BibTeX{{\rm B\kern-.05em{\sc i\kern-.025em b}\kern-.08em
    T\kern-.1667em\lower.7ex\hbox{E}\kern-.125emX}}
\begin{document}

\title{Using Causal Trees to Estimate Personalized Task Difficulty in Post-Stroke Individuals}

\author{Nathaniel Dennler, Stefanos Nikolaidis, \& Maja Matari\'c
\thanks{All authors are from the Computer Science Department at the University of Southern California, Los Angeles, USA. 
    {\tt\small \{dennler, mataric\}@usc.edu; stefanosnikolaidis@gmail.com}}%
}

\maketitle

\begin{abstract}
Adaptive training programs are crucial for recovery post stroke. However, developing programs that automatically adapt depends on quantifying how difficult a task is for a specific individual at a particular stage of their recovery. In this work, we propose a method that automatically generates regions of different task difficulty levels based on an individual's performance. We show that this technique explains the variance in user performance for a reaching task better than previous approaches to estimating task difficulty.
\end{abstract}

\section{Introduction}
To facilitate motor learning in people with mobility loss, physical therapists administer programs that consider the individual's performance as they recover, and adapt exercises to match task difficulty to individual \cite{guadagnoli2004challenge}. Physical therapists must take the time to evaluate and adjust the difficulty of assigned exercised based on user performance. Robotic companions could assist the process by providing such evaluations and enabling more people access to physical therapy in order to regain mobility through more regular practice.

Determining task difficulty for individuals with limited mobility is complex, as the specific limitations are highly variable and depend on many contextual factors. For example, reaching to the left may be more difficult for an individual than reaching to the right, even for equidistant points.

In this work, we propose a method based on using causal trees to automatically learn regions of similar task difficulty based on task-related features. These regions can be used to communicate what aspects of a task are difficult for a particular individual, informing physical therapists how to shape and personalize rehabilitation programs for each individual. We use a reaching task to evaluate our method, a common task for evaluating performance characteristics in stroke \cite{han2013quantifying}.

\begin{figure*}[t!]
    \centering
     \begin{subfigure} {.35\linewidth}
      \includegraphics[width=\linewidth]{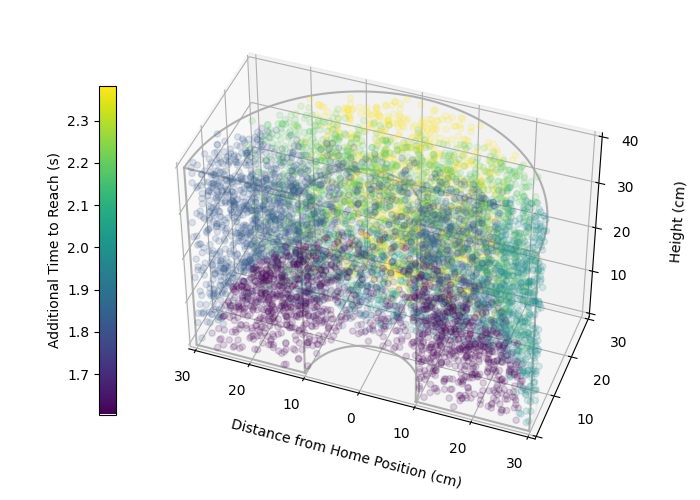}
        \caption{PID 25 map of difficulties}
    \end{subfigure}
    \begin{subfigure} {.35\linewidth}
      \includegraphics[width=\linewidth]{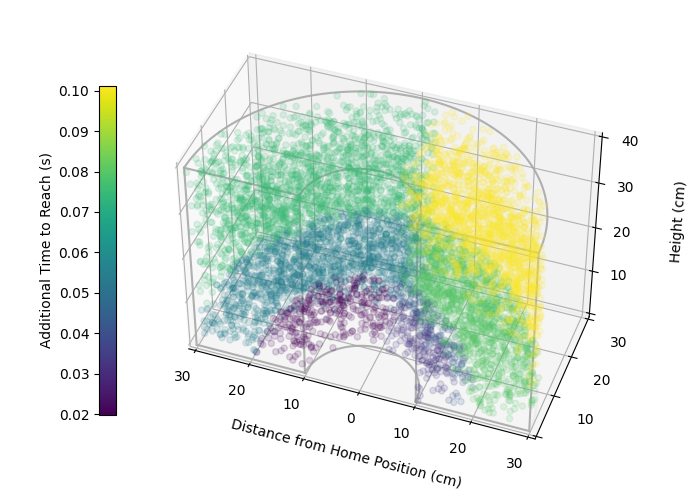}
       \caption{PID 27 map of difficulties}
     \end{subfigure}
     \caption{Maps of difficulties for two different participants. PID 25 was left-affected and had lower overall Fugyl-Meyer score (47 out of 65). PID 25 exhibited more difficulty reaching far in front. PID 27 is right-affected and had a higher Fugyl-Meyer Score (62 out of 65). PID 27 exhibited more difficulty reaching higher points on their right side.}
     \label{fig:personalized_difficulty}
\end{figure*}

\section{Related Work}

Previous work on quantifying task difficulty has generally taken one of two approaches. One is to assess the difficulty associated with several specific motor movements in isolation \cite{woodbury2016matching, metzger2014assessment}, while the other uses concepts from control theory to assist people throughout entire tasks as needed \cite{asl2017assist,basteris2013adaptive}.

Assessing task difficulty through specific movements in isolation can create a very detailed understanding of personal differences when performing movements such as pinching, elbow flexion, and arm supination \cite{woodbury2016matching}. When individuals use these types of movements in concert when doing more holistic activities, like reaching, the combination of movements changes the task difficulty.

For control theory approaches, robots can adapt in real time to an individual to account for their level of mobility and other aspects that affect performance, such as fatigue \cite{basteris2013adaptive}. Often, these approaches rely on exo-skeleton-type assistance \cite{asl2017assist}. However, the robots can be encumbering or scary to some individuals who prefer to interact with robots that do not exert forces on them to perform certain tasks that reflect daily life activities. Additionally, controls-based approaches require continuous measurements of performance error that may not be possible for daily tasks like cup-stacking, which can be best evaluated only after completion.

In this work we aim to address the shortcomings of existing approaches by developing a technique that estimates how difficult a task is based on features of the task. Our technique applies to holistic tasks and does not require the robot to exert forces on the user.

\section{Problem Statement}

We draw upon work from estimating heterogenous treatment effects from covariates (e.g., how a drug may have different effects for someone based on their medical characteristics) \cite{wager2018estimation}. We adapt this approach to the context of task difficulty estimation by making the observation that there are two components to evaluating task difficulty in individuals: 1) the component of difficulty inherent to the task (i.e., how long it takes a healthy control population on average to complete the task), and 2) the component that comes from individual differences in movement (i.e, how much shorter or longer than the average does it take for an individual to complete the task). We aim to quantify the second component in this work.

Formally we are given a space, $X$, that defines all possible tasks. We are also given a continuous function that represents success in a task, $Y$. We define the participant indicator as $W \in \{0,1\}$, where 0 indicates that the X,Y values are from the neurotypical baseline, and 1 indicates the values are from a the individual we are finding personal differences for. Using the potential outcomes framework \cite{rubin1974estimating}, we define the effect of individual task difficulty for task $x_i$ as:

\begin{equation}
    \tau(x) = \mathbb{E} \left[ Y^{(1)}_i - Y^{(0)}_i | X=x_i \right]
\end{equation}

Estimating $\tau(x)$ directly is difficult because $Y_i$ is typically noisy due to unmodelled effects of the task. Thus, previous works have found that creating a decision tree to estimate the causal effect can increase the power of the predictions by averaging over several points as defined by the leaves of the tree. The decision tree learns a function $L(x)$ that assigns the parameterization of the task to different leaves. The estimated personalized difficulty is then expressed for each leaf $L$ as:

\begin{equation}
    \hat{\tau}(x) = \frac{1}{| \{ x^{(1)}_i \in L \} |} \sum_{x^{(1)}_i \in L} Y^{(1)}_i - \frac{1}{|\{ x^{(0)}_i \in L \} |} \sum_{x^{(0)}_i \in L} Y^{(0)}_i
\end{equation}

The tree is learned through the techniques outlined by Wagner and Athey \cite{wager2018estimation} using the econML package \cite{econml2019econml}.

\section{Dataset for Evaluating Approach}
We used a dataset we previously collected from post-stroke participants perforiming a reaching task with a robot \cite{dennler2023metric}. The participants were assessed by a physical therapist to determine their Fugyl-Meyer scores before the study.

During the study, participants placed their hands on a ``home position" directly in front of them. A robot arm held a button that moved within the workspace of the participant. The workspace was defined as a semicircle region that extended 30cm out from the home position in the XY plane, and 40cm high in the z direction. The participants reached to the button as quickly as possible when the light on the button turned on. We measured the time to press the button following the cue. For this task, the parameterization of the task space $X$ is x,y,z, and distance, and the success metric $Y$ is time to reach using the individual's more-affected side.

\section{Preliminary Results}

\subsection{Estimating Personalized Difficulty}
We compared the results from the causal tree which jointly learns L(x) from both treatment and control data for each participant with the common alternative of separately learning functions for $Y^{(1)}$ and $Y^{(0)}$ and subtracting the estimated individual's completion time and control completion times for all parameterizations of the task. The ground truth values were calculated from held out reaching datapoints. Table \ref{table:results} reports the $r^2$ goodness of fit values for all held out points and across all participants, averaged over 10 runs. We show that causal trees significantly outperform all baselines when estimating personalized task difficulty. 

\begin{table}
\caption{Performance of Predicting Personalized Difficulty}
\label{table:results}

\begin{tabular}{ rccc } 
\hline
Model & avg. $r^2$ & std. err. $r^2$ & \textit{p-value}\\
\hline
Causal Tree & \textbf{.656} & .007 & -- \\
Random Forest & .635 & .006 & .036\\
Neural Network & .634 & .005 & .023\\ 
Support Vector Machine & .598 & .010 & $<.001$\\
5-Nearest Neighbors & .572 & .007& $<.001$\\
Decision Tree & .588 & .007 & $<.001$\\
\hline
\end{tabular}
\end{table}

\subsection{Visualizing Personalized Difficulty}
Figure \ref{fig:personalized_difficulty} illustrates that the representations of personalized task difficulty provide interpretable feedback for both individuals and clinicians. This feedback shows the areas where the individual performs differently than the control participants. We also note that some areas that have similar levels of difficulty are disjoint, which allows the robot to exhibit more variation in the task than a controls-based approach that would only slightly move the location of the button, leading to a less varied interaction.

\section{Discussion and Conclusions}

In this work we outlined a method to identify areas where users have different levels of difficulty from baseline performance on a task that is parameterized by a set of features. Our method provides better estimates of actual task performance that a robot can use to adapt task difficulty levels for each user, and it provides an interpretable way for users and clinicians to understand what tasks are most important to practice. The described method can be readily applied to different robotic rehabilitation tasks (e.g., pushing or grasping tasks). Future works can evaluate how this method can be used within an interaction to select appropriate levels of difficulty for a specific user.

\bibliographystyle{IEEEtran}
\bibliography{references}

\end{document}